\pdfoutput=1

\documentclass[11pt]{article}
\usepackage[preprint]{acl}

\usepackage{times}
\usepackage{latexsym}
\usepackage{setspace}
\usepackage{multirow}
\usepackage{floatrow}
\usepackage{amsmath}
\usepackage{graphicx}
\usepackage{enumitem}
\usepackage{booktabs}
\usepackage{color}
\usepackage{cleveref}
\usepackage{CJK}
\usepackage{url}

\usepackage[T1]{fontenc}

\usepackage[utf8]{inputenc}
\usepackage{microtype}

\usepackage{inconsolata}

\usepackage{graphicx}

\title{Classic4Children: Adapting Chinese Literary Classics for Children with Large Language Model}

\newcommand{\ie}{\emph{i.e.}}
\newcommand{\eg}{\emph{e.g.}}

\author{
 \textbf{Jiali Chen\textsuperscript{1,2}},
 \textbf{Xusen Hei\textsuperscript{1,2}},
 \textbf{Yuqi Xue\textsuperscript{1,2}},
 \textbf{Zihan Wu\textsuperscript{1,2}},
 \textbf{Jiayuan Xie\textsuperscript{3}},
 \textbf{Yi Cai\textsuperscript{1,2}$^*$}
\\
 \textsuperscript{1}Key Laboratory of Big Data and Intelligent Robot
\\
(South China University of Technology) Ministry of Education, \\
 \textsuperscript{2}School of Software Engineering, South China University of Technology, \\
 \textsuperscript{3}The Hong Kong Polytechnic
University
\\
 \small{
   \textbf{$^*$Correspondence:} \href{mailto:ycai@scut.edu.cn}{ycai@scut.edu.cn}
 }
}

\begin{document}
\maketitle
\begin{abstract}
Chinese literary classics hold significant cultural and educational value, offering deep insights into morality, history, and human nature. These works often include classical Chinese and complex narratives, making them difficult for children to read.
To bridge this gap, we introduce a child-friendly literary adaptation (CLA) task to adapt the Chinese literary classic into engaging and accessible text for children.
However, recent large language models (LLMs) overlook children's reading preferences (\ie, vivid character portrayals, concise narrative structures, and appropriate readability), which poses challenges in CLA. 
In this paper, we propose a method called InstructChild, which augments the LLM with these preferences for adaptation.
Specifically, we first obtain the characters' personalities and narrative structure as additional information for fine-grained instruction tuning. 
Then, we devise a readability metric as the reward to align the LLM with the children's reading level.
Finally, a lookahead decoding strategy is applied to improve the readability of the generated text during inference.
To support the evaluation of CLA task, we construct the Classic4Children dataset, which comprises both the original and child-friendly versions of the Four Great Classical Novels of Chinese literature.
Experimental results show that our InstructChild significantly improves automatic and human evaluation performance.
\end{abstract}

\section{Introduction}
The Chinese literary classics portray iconic characters and their stories to convey significant cultural and educational value \cite{cn-classic-2,cn-classic-1}. 
These works are not only a vital part of China's rich literary heritage but also play a crucial role in shaping moral values and cultural understanding, especially for young learners. 
Given their importance, these classics are a fundamental part of the Chinese educational curriculum, and many children are required to engage with these texts during their schooling years \cite{edu-curriculum-1}.
However, these works are frequently written in classical Chinese and involve complex plots and themes, presenting significant challenges for children to read.
Traditionally, many writers make much effort to manually transform these complex literary classics into child-friendly versions \cite{ada-1,ada-2}, which is a time-consuming and labor-intensive process.
\begin{figure}[]
  \centering
  \includegraphics[scale=0.314]{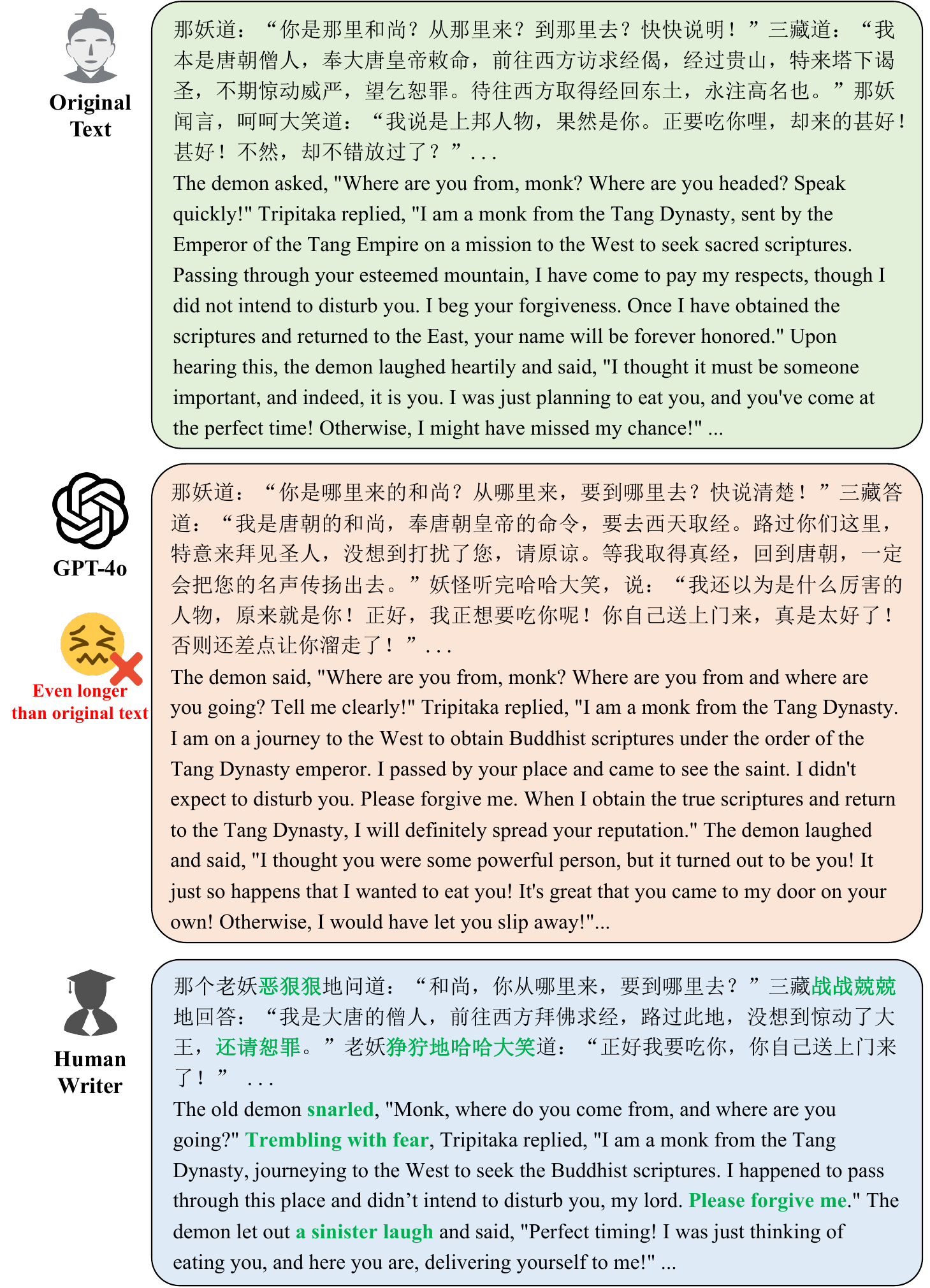}
  \caption{The sample from our Classic4Children dataset. 
  We use a one-shot human-written example in Table \ref{prompt_llms} to query GPT-4o.
  The \textcolor[RGB]{0, 176, 80}{green text} emphasizes the character's personality traits.}
\label{intro_case}
\end{figure}
Therefore, we propose the child-friendly literary adaptation (CLA) task, which aims to automatically make content accessible and engaging for children.

Promisingly, the recent advanced large language models (LLMs) have performed impressively across various natural language processing tasks, including text style transfer  \cite{llm-tst-0,llm-tst-1} and text simplification \cite{child-ts,llm-ts}. 
Nevertheless, such methods simply modify specific stylistic elements (\eg, sentiment, formality, author-style and lexicon) within sentences, which overlook the importance of children's reading preferences, leading to poor performance in the CLA task. 
Moreover, it is well known that LLMs can capture language patterns from provided examples through in-context learning \cite{icl}. We utilize a carefully crafted prompt with a one-shot human-written example to query GPT-4o \cite{gpt4}, exploring whether children's reading preferences can be captured by GPT-4o, as shown in Fig. \ref{intro_case}.
Surprisingly, it also fails to produce text similar to human-written content, instead generating an overly lengthy translation of classical Chinese without simplification, which is unsuitable for children.
These findings underscore the limitations of current LLMs in effectively adapting literary classics for children. 

Scrutinizing the adapted child-friendly text by human writers, we identify three key children's reading preferences that are crucial for effective adaptation.
\textbf{i)} Considering that the literary classic often contains numerous characters, vividly portraying each character’s personality can help children better remember and distinguish them. 
As shown in Fig. \ref{intro_case}, the green-highlighted words in the adaptation emphasize the different personalities of Tripitaka and the Demon. 
For instance, Tripitak presents a timid and respectful nature by his trembling response, showcasing his fear and humility in the face of danger. 
\textbf{ii)} Concise narrative structure, rather than overly detailed or complex plots, is effective in sustaining children's interest in reading. 
As shown in Fig. \ref{intro_case},
the adapted text simplifies the dialogue while maintaining the core storyline.
Specifically, several unnecessary details, like ``sent by the Emperor'' and ``obtained the scriptures'' are removed in Tripitaka's explanation, as they introduce background information and future outcomes that may confuse children.
\textbf{iii)} Besides, the adaptation also take into account the children's reading level to ensure appropriate readability, facilitating easier comprehension and engagement with the content \cite{child-red}.

In this paper, we propose the InstructChild, a method to effectively adapt Chinese literary classics for children with the LLM.
It consists of three key techniques: 
First, we apply \textbf{fine-grained instruction tuning}, which incorporates personality information and narrative structure for LLM to generate text that emphasizes character traits and follows a concise narrative.
Specifically, the personality is assessed based on the Big Five Personality Traits (BFPT) \cite{bfpt} theory and the narrative structure is presented by the entity-relation triplets of the original text.
Next, we design a Chinese readability metric to guide the \textbf{refinement} process. 
It ensures the LLM generates adapted text with easier-to-understand sentences for children.
Finally, inspired by \cite{scott}, during inference, the \textbf{lookahead decoding strategy} considers the impact of potential subsequent tokens based on the readability metric for current token selection.
In addition, we collect both original and adapted versions of the Four Great Classical Novels of Chinese literature to construct the Classic4Children dataset for evaluation.

Our main contributions are: 
\textbf{i)} We highlight children's reading preferences (\ie, vivid character portrayals, concise narrative structure and appropriate readability) are essential in adapting Chinese literary classics for children (CLA). We also find that state-of-the-art LLMs struggle to capture these preferences in their adaptations.
\textbf{ii)} We propose the InstructChild for the CLA task, which effectively incorporates these preferences with the LLM, using fine-grained instruction tuning, refinement with readability metric and lookahead decoding strategy.
\textbf{iii)} We construct Classic4Children dataset from the Four Great Classical Novels of Chinese literature for evaluation. 
Our InstructChild achieves significant performance gains over the existing LLMs.

To facilitate further research, we make the code and dataset available at \url{https://github.com/Gary-code/Classic4Children}.

\section{Related Work}
\subsection{Text Style Transfer}
The objective of text style transfer (TST) is to endow text with a different style (\eg, positive $\rightarrow$ negative) while preserving its semantic content. 
The traditional paradigm explicitly divides text into content and style information and then employs a target style for desired text generation ~\cite{yl-sen-con,storytrans,tst-sc2}.
Specifically, \citet{storytrans} address the task of author-style transfer and implement content-style disentanglement and stylization at the discourse level.
\citet{tst-sc2} develop a multilayer Joint Style-Content Weighed (JSCW) module along with a style consistency loss to ensure both content preservation and consistent style across generated sentences.
Recently, LLMs have shown promising results on TST through fine-tuning ~\cite{tst-ft-1, tst-ft-2}, in-context learning \cite{tst-icl-1,tst-icl-3} and promt-based editing \cite{p-r,tst-pd-2}.
However, previous methods primarily transfer the text styles related to sentiment and formality.
This study adapts Chinese literary classics into a child-friendly style, emphasizing vivid character descriptions and concise narrative structure tailored to children’s reading levels.
\begin{figure*}[]
  \centering
  \includegraphics[scale=0.23]{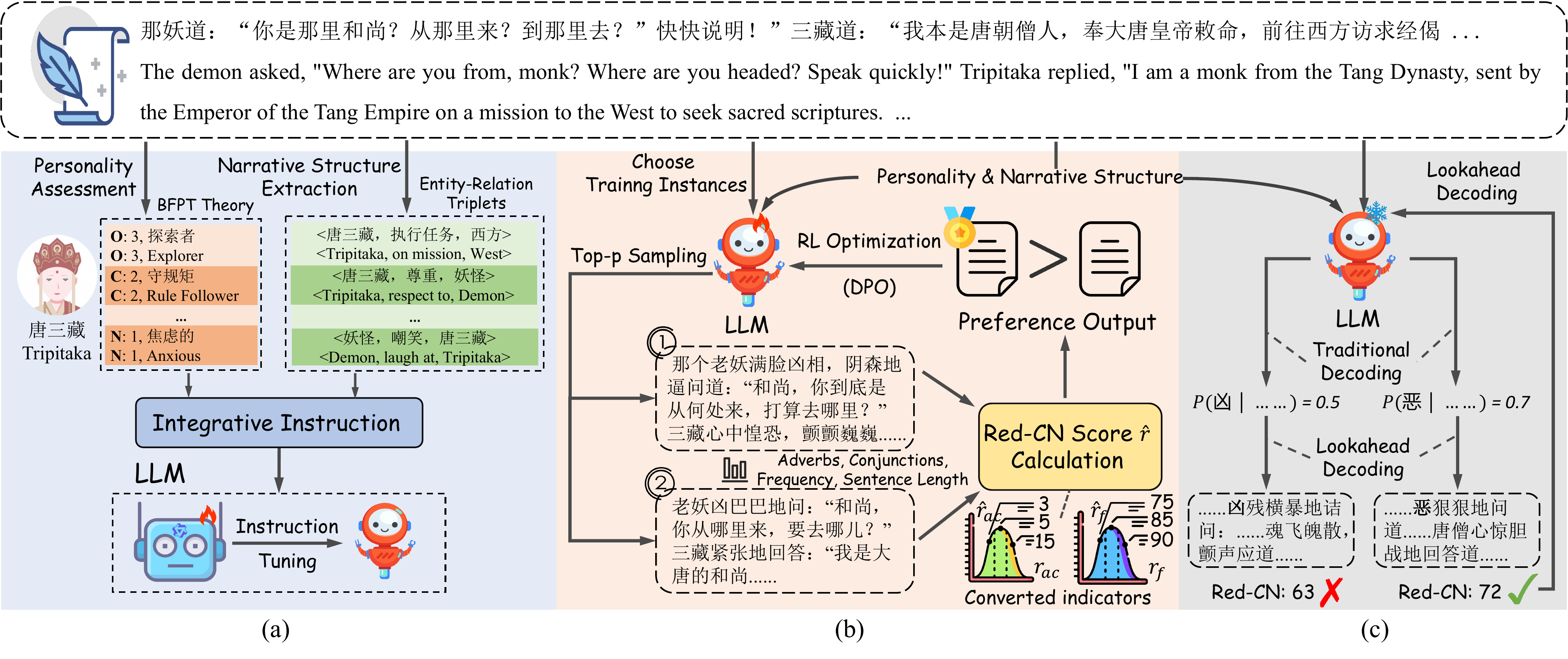}
  \caption{Overview of InstructChild.
  (a) fine-grained instruction tuning, which incorporates the characters' personalities and narrative structure to fine-tune the large language model (LLM) with LoRA.
  (b) refinement with reinforcement learning, which considers a readability metric (\ie, Red-CN) as a reward to further optimize 
  the LLM to align with the children's reading level.
  (c) lookahead decoding strategy, which extends the traditional decoding strategy with the readability score during inference. } 
  \label{framework}
\end{figure*}

\subsection{Text Simplification}
Text simplification aims to reduce the complexity of the text, 
which can be categorized into two main branches (\ie, editing operations and lexical-syntactic rules).
Specifically, 
the editing operations simplify the text through various editing techniques, such as replacing difficult words and reordering sentence components \cite{edit-ts-1,edit-ts-2}. 
Moreover, some works introduce the lexical and syntactic rules for simplification \cite{ts-lsbert,child-ts}.
For instance, \citet{child-ts} incorporate the LLM with the lexical simplification model to simplify the children's story.
In contrast to text simplification, which aims to simplify vocabulary and syntax, Chinese literary classics often feature classical language that is challenging for children to understand. 
Our approach involves clearly explaining complex expressions while maintaining children's engagement through vivid character descriptions and an age-appropriate narrative structure.

\section{Methodology}
In this section, we introduce InstructChild, a large language model (LLM) based method to facilitate the adaptation of Chinese literary classics into a child-friendly style. 
This method bridges the gap between the complexity of the original text and the cognitive abilities of children with fine-grained instruction tuning, refinement and a lookahead decoding strategy.
Fig. \ref{framework} illustrates the proposed InstructChild, and the details of our method are elaborated in the following sections.

\subsection{Fine-grained Instruction Tuning}
Inspired by previous studies \cite{scott,instruct-gpt,yl-logic} that successfully apply well-designed prompts as guidance to generate text with desired attributes, 
we develop instruction incorporating personality and narrative structure to fine-tune the LLM (\ie, Qwen2-7B-Instruct ~\cite{qwen2}), aiming to adapt text with vivid character portrayals and concise narrative structure.
Specifically, they are prepended to input text as the instruction for fine-tuning.

\subsubsection{Personality Assessment}
In the field of psychology, the Big Five Personality Traits (BFPT) \cite{bfpt} categorizes personality into five dimensions: openness, conscientiousness, extraversion, agreeableness, and neuroticism.
It provides a robust framework for assessing character personality.
Technically, 
we first identify the character names in the original text and filter out the top 50 characters by frequency of occurrence for each literature.
Given that large language models (LLMs) have demonstrated the ability to effectively capture the personalities of characters \cite{llm-role, role-cha},
we prompt GPT-4o \cite{gpt4} to obtain the personality dimension scores and brief descriptions of characters in the original text. The prompt is shown in the Table \ref{beft_prompt}.
Specifically, the score for each personality dimension ranges from 1 to 5, with higher values indicating a stronger presence of that specific dimension.
Consequently, we obtain the personality information for each character based on these dimensions with the corresponding score and brief description, as shown in Fig. \ref{framework}(a).

\subsubsection{Narrative Structure Extraction}
In addition to character personality, the concise narrative structure also distinguishes adapted child-friendly style text from the original literary classic \cite{ada-1}.
It can be constructed from the entities and their relationships in the text \cite{llm-cr,kb-vqa-ex}. 
Therefore, we also employ the GPT-4o to identify entities and relationships as supplementary information for further fine-tuning. The prompt is shown in the Table \ref{ent-rel_prompt}.
 These triplets can be integrated into the instructions to guide the LLM in focusing on important narrative elements.

\subsubsection{Integrative Instruction for LLM}
The integrative instruction comprises the characters' personalities, narrative structure (\ie, entity-relation triplets) and the original literary texts. 
Considering that each input literary text typically contains only a limited number of characters, we first identify the character names and then retrieve the corresponding personality information of the character.
An exemplar instruction for fine-tuning the large language model is shown in Table \ref{instruction}. 
The integrative instruction is directly fed into the frozen large language model (\ie, Qwen2-7B-Instruct) with learnable LoRA layers for child-friendly style text adaptation. The language modeling loss can be formulated as:
\begin{equation}
    \mathcal{L}_{lan}=-\sum_{t=1}^T \log p\left(y_t \mid {y}_{<t}, \mathrm{Ins}\right),
    \label{l_lan}
\end{equation}
where $\log p$ is the negative log-likehood, $\mathrm{Ins}$ is the integrative instruction, and ${y}_{<t}$ represents the words before the $t$-th word.

\subsection{Refinement with Readability Metric}
After fine-grained instruction fine-tuning, the LLM can effectively emphasize character personalities and the narration of a concise storyline.
However, the instruction tuning phase does not include an explicit mechanism to verify whether the generated text aligns with the children's reading level.
Instead, it relies solely on token-wise gradient updates through teacher forcing based on reference texts.
Therefore, we design the readability metric tailored for Chinese child-adapted literary classics, namely Red-CN, to assist LLM in generating text with the desired readability.
Existing studies ~\cite{instruct-gpt,instructscore},
have demonstrated that 
supplementing initial instruction tuning with a subsequent reinforcement learning phase can be beneficial, 
where the model is further refined by the reward function. 
For our implementation, we utilize the designed readability metric Red-CN as a reward for refinement.

\subsubsection{Readability Metric as Reward}
We design the readability metric Red-CN, which takes into account the complexity of Chinese sentences.
Inspired by \cite{child-ts,child-red}, we first assess the suitability of Chinese characters and syntax. 
Specifically, the metric is determined by the proportion of adverbs and conjunctions $r_{ac}$ and the frequency statistics of the character $r_f$ based on the collection \cite{cn-sta} in each sentence.
We analyze expert-adapted child-friendly literary classics and find that these two indicators typically cluster around values of 5 and 85 per sentence, respectively. 
It suggests that excessively low or high values of these indicators are not conducive to optimal reading levels for children.
We assign the maximum value of 1.0 when any indicator matches the desired value when each indicator matches its corresponding target value previously mentioned, with the reward decreasing exponentially toward 0 as it deviates from the target values. 
Thus, we convert these two indicators with the normalized Gaussian distribution $\mathcal{F}(\cdot)$ centered at their target values:
\begin{equation}
    \mathcal{F} = \mathcal{N}(\overline{r}, \delta^2),
\label{eq_gus}
\end{equation}
\begin{equation}
\hat{r}_{ac} = \mathcal{F}(r_{ac}),  \  \hat{r}_{f} = \mathcal{F}(r_f),
\end{equation}
where $\overline{r}$ is the corresponding target value, $\delta$ is the standard deviation. $\hat{r}_{ac}$ and $\hat{r}_f$ are the converted indicators. 
The rationale behind introducing the Gaussian distribution is to ensure that the reward decreases nonlinearly, in the sense that minor deviations from the target readability lead to slight reductions, whereas more significant deviations result in increasingly larger penalties. More details are shown in the Appendix \ref{app-met-1} and \ref{app-met-2}.

Furthermore, \citet{less-word} suggest that reducing the number of characters contributes to more effective adaptations of literary classics for children. We use the token length of the adapted text as an additional indicator to prevent generating overly verbose text. 
This indicator is also normalized to the range of [0, 1], using its proportional value relative to the original input text, denoted as $\hat{r}_t$, where higher values represent fewer tokens:
\begin{equation}
    \hat{r}_{t} = \max\left(0, 1 - \frac{\mathrm{output\_len}}{\mathrm{input\_len}}\right),
\end{equation}
where $\mathrm{input\_len}$ and $\mathrm{output\_len}$ represent the length of the original text and the generated text, respectively.
Finally, following the emphasis by \citet{cn-sta} on the strong correlation between character frequency and readability in Chinese, we assign the specific weights to obtain the overall readability metric (\ie, Red-CN), as the reward score $\hat{r}$:
\begin{equation}
    \hat{r} = 0.3\hat{r}_{ac} + 0.4\hat{r}_{f} + 0.3\hat{r}_t.
    \label{eq_reward}
\end{equation}

We also report correlation coefficients in Table \ref{coff} of Appendix \ref{sec-coff}, showing good correlations between Red-CN and human evaluation metrics.

\subsubsection{DPO}
The reinforcement learning with the readability metric is used to ensure the LLM generates text aligned with the children's reading level.
Specifically, we randomly choose 1,000 instances from the Classic4Children training data and use the top-p sampling method to produce $K$ outputs for each sample, represented as $\{\hat{Y}_1, \hat{Y}_2, ..., \hat{Y}_K\}$.
Next, we form them as rank pairs based on the readability metric score, ignoring those with a score difference less than 3.
The direct preference optimization (DPO) \cite{dpo} is employed as a reinforcement learning policy to further optimize the LLM with learnable LoRA layers. 

\subsection{Lookahead Decoding}
Getting inspiration from \cite{scott}, 
we devise a lookahead decoding strategy during inference to maximize the LLM's ability to generate text with high readability scores, which extends the traditional strategy with the readability metric, as illustrated in Fig. \ref{framework}(c). 
The core idea involves forecasting potential subsequent tokens and then adjusting the selection process toward higher readability scores.
Specifically, the LLM first generates $L$ candidate samples at the $t$-th decoding step. Each sample starts at the $(t-1)$-th step and continually generates $n$ subsequent tokens, which can be denoted as $\hat{y}_{<t-1+n}$.
Then, we calculate the readability scores of these candidate samples. The calculation of each sample can be formulated as:
\begin{equation}
    \mathcal{G}(\hat{y}_{<t-1+n}) = \hat{r},
\end{equation}
where 
$\mathcal{G}(\cdot)$ denotes an supplementary guidance function during decoding. 
Consequently, the selection criterion for the $t$-th token can be mathematically expressed as:
\begin{equation}
    f(y_t) =  \log p(y_t \mid y_{<t}, \mathrm{Ins}) + \lambda \max _{L} \mathcal{G}\left(\hat{y}_{<t-1+n}\right),
    \label{eq_select}
\end{equation}
where $\lambda$ is the hyperparameter that adjusts the influence of readability on the token selection process.

\section{Experiment}

\subsection{Classic4Children Dataset}\label{sec-dataset}
In this paper, we construct Classic4Children, a Chinese child-friendly literary adaptation dataset based on the Four Great Classical Novels of Chinese literature
(\ie, Journey to the West, Romance of the Three Kingdoms, Water Margin, Dream of the Red Chamber).
Initially, we collect widely recognized versions of the original texts and their corresponding child-adapted editions\footnotemark\footnotetext{https://5000yan.com/}\footnotemark\footnotetext{http://www.bph.com.cn/shaoer.html?\_isa=1}.
Following \cite{storytrans}, 
since there are too many tokens in each chapter of the literary classic, we manually annotate the corresponding paragraph fragments as training samples, with each chapter being divided into multiple fragments. 
Additionally, these fragments generally span several paragraphs.
Finally, we manually match these fragments from both versions as training pairs. 
The Classic4Children dataset consists of 2,686 and 300 samples for training and testing, respectively.
More details of the dataset construction are described in the Appendix \ref{app-dataset}.

\subsection{Implementation Details}
We train our InstructChild on a Tesla A40 48GB GPU card and it is initialized with Qwen2-7B-Instruct\footnotemark\footnotetext{https://huggingface.co/Qwen/Qwen2-7B-Instruct}.
We insert the LoRA layers into each self-attention layer, setting the rank to 8 for optimization.
Specifically, the batch size is 24 and we train 3 epochs during the fine-grained instruction tuning. 
The desired values (\ie, the expectation) of the Gaussian distribution in Eq. \ref{eq_gus} are 5 and 85 for $\hat{r}_{ac}$ and $\hat{r}_f$, respectively. 
After experimenting with different values for the standard deviation $\delta$, we set it at half the expectation (\ie, 2.5 and 42.5, respectively).
This choice allows the mapping values to be concentrated as closely as possible around the target values.
Before the refinement, we randomly sample 1,000 instances from the training set and employ top-p sampling strategy, where the temperature is set to 0.8 and $p$=0.9 to generate four candidate samples for each input. We maintain those with a score difference greater than 3 for direct preference optimization.
For the lookahead decoding strategy, the LLM generates $L$=5 candidate samples with $n$=20 subsequent tokens at each decoding step. The hyperparameter $\lambda$ for the token selection in Eq. \ref{eq_select} is 1.

\subsection{Baselines and Ablation Models}

\subsubsection{Baseline}
To verify the superiority of our InstructChild, we compare it with three types of baselines.
\textbf{i)} Closed-source large language models (LLMs) with large-scale parameters (\textbf{API-based}): GPT-4o \cite{gpt4}, GLM-4 \cite{glm4}, QWen2.5 \cite{qwen2}, and EAPMT \cite{eapmt}. 
We access these models via API interfaces, which exhibit strong reasoning capabilities across various natural language processing tasks. Specifically, we also ask them to focus on the characters personality and concise narrative structure with a one-shot example.
In particular, EAPMT first generates a detailed explanation for each input sentence, encompassing both the literal content and deeper meanings.
Then it paraphrases sentences into a child-friendly style format based on the explanation.
The prompts for these LLMs is shown in Table \ref{prompt_llms} and Table \ref{prompt_EAPMT}.
\textbf{ii)} Text style transfer (TST) models (\textbf{TST-based}): P\&R \cite{p-r}, StoryTrans \cite{storytrans}, and ParaGuide \cite{paraguide}. 
Specifically, P\&R is a training-free few-shot TST model and we utilize the Qwen2 as the backbone. 
We also expand the Chinese vocabulary for diffusion-based Paraguide and re-trained it for our task.
\textbf{iii)} Open-source large language models with relatively moderate parameters (\textbf{FT-based}): Llama2-13B \cite{Llama2}, Llama3-8B \cite{Llama3}, Qwen2-7B \cite{qwen2} and GLM4-9B \cite{glm4}. 
We utilize LoRA \cite{lora} technique to fine-tune these LLMs.
Notably, Llama models have been extended with the Chinese vocabulary and further pre-trained on Chinese instruction \cite{Llama-cn}.

\subsubsection{Ablation Models} 
We conduct ablation experiments to evaluate the effectiveness of various components within our InstructChild method. 
\textbf{InstructChild w/o Per} and \textbf{InstructChild w/o Nat}: InstructChild without personality information and narrative structure, respectively.
\textbf{InstructChild w/o Ref}: InstructChild without refinement stage by reinforcement learning. It does not implement the lookahead decoding strategy for generation, as it is closely tied to the Red-CN metric used during refinement.
\textbf{InstructChild w/o Look}: InstructChild without lookahead decoding strategy during inference.

\subsection{Evaluation Metric}
\subsubsection{Automatic Evaluation Metrics}
Following previous studies \cite{storytrans}, we evaluate the performance with BLEU-(1 to 2) \cite{bleu} and BERTScore \cite{bert-score} metrics. 
Specifically, we report the recall (BS-R), precision (BS-P) and F1 score (BS-F1) of the BERTScore. 
They are commonly used for measuring the lexical and semantic similarity in text generation.
Moreover, we also use the readability scores in Eq. \ref{eq_reward} as an additional metric (\ie, Red-CN) to assess whether the generated text aligns with children's reading levels (higher is better).

\subsubsection{Human Evaluation Criteria}
We also conduct human evaluation for baselines (\ie, GPT-4o, StoryTrans and GLM4-9B) and our InstructChild. 
Specifically, we randomly select 200 samples from the test set for evaluation and invite five volunteers with good educational backgrounds in Chinese to assess the following criteria: 
Fluency (\textbf{Flu}) mainly reflects the grammatical correctness and fluency of the generated text. 
Content Preservation (\textbf{CP}) refers to the conservation of the core storyline and coherence of the narrative.
Character Clarity (\textbf{CC}) measures the clarity and emphasis of character descriptions. 
Narrative Efficiency (\textbf{NE}) indicates the conciseness of the story narration with less redundant plots. 
Each criterion is scored on a scale from 0 to 2, with higher scores indicating better performance.

\subsection{Results and Analysis}
\begin{table*}[]
\centering
\begin{spacing}{0.9}
\resizebox{1.\columnwidth}{!}{
\begin{tabular}{c|c|c|c|c|c|c|c}
\toprule[1pt]
\textbf{Method}                             & \textbf{Type}                   & \textbf{BLEU-1}       & \textbf{BLEU-2}       & \textbf{BS-P}         & \textbf{BS-R}         & \textbf{BS-F1}        & \textbf{Red-CN}  \\  \midrule
GPT-4o  \cite{gpt4}                                     & \multirow{4}{*}{API-based}   & 20.91                 & 10.42                 & 70.59                 & 65.96                 & 68.04                 & 68.19                \\ \cline{1-1} \cline{3-8} 
GLM-4  \cite{glm4}                                     &                             & 20.57                 & 10.54                 & 71.44                 & 65.04                 & 67.97                 & 61.95                \\ \cline{1-1} \cline{3-8} 
QWen2.5  \cite{qwen2}                                   &                             & 20.30                  & 9.30                   & 69.84                 & 64.92                 & 67.17                 & 63.09                \\ \cline{1-1} \cline{3-8} 
EAPMT   \cite{eapmt}                                    &                             & 18.26                 & 8.91                  & 71.42                 & 63.09                 & 66.89                 & 57.47                \\ \hline
P\&R    \cite{p-r}                                    & \multirow{3}{*}{TST-based} &  8.60                 &   4.12               &   57.94                    &    59.32                   &   58.62                    &  56.73                \\ \cline{1-1} \cline{3-8} 
StoryTrans \cite{storytrans}                                 &                             & 7.59                  & 4.41                  & 59.19                 & 60.32                 & 59.65                 & 69.74                \\ \cline{1-1} \cline{3-8} 
Paraguide  \cite{paraguide}                                 &                             & 13.88                 & 4.64                  & 60.43                 & 61.24                 & 60.72                 & 71.87                \\ \hline
Llama2-13B \cite{Llama2}                                 & \multirow{4}{*}{FT-based}   & 13.78                 & 8.47                  & \textbf{73.86}                 & 60.72                 & 66.51                 & 58.89                \\ \cline{1-1} \cline{3-8} 
Llama3-8B  \cite{Llama3}                                 &                             & 8.97                  & 5.45                  & 73.66                 & 58.60                  & 65.09                 & 50.06                \\ \cline{1-1} \cline{3-8} 
QWen2-7B  \cite{qwen2}                                 &                             & 17.63                  & 8.09                  & 66.04                 & 68.12                  & 67.07                 & 67.31                \\ \cline{1-1} \cline{3-8} 
GLM4-9B   \cite{glm4}                                  &                             & 16.87                 & 7.16                  & 65.80                  & 67.42                 & 66.44                 & 67.54                \\ \hline
InstructChild w/o Per                          & \multirow{5}{*}{-}          & 20.07                 & 9.63                 & 67.87                 & 68.32                 & 68.10                 & 71.85                \\ \cline{1-1} \cline{3-8} 
InstructChild w/o Nat                    &                             & 21.46                 & 10.93                 & 67.96                 & 69.03                 & 68.21                 & 71.56                \\ \cline{1-1} \cline{3-8} 
InstructChild w/o Ref                            &                             & 21.37                 & 11.98                  & 69.55                 & 70.26                 & 69.36                 & 67.88                \\ \cline{1-1} \cline{3-8} 
InstructChild w/o Look                            &                             & 22.45                      & 12.89                      & 69.72                      & 70.16                      & 69.94                      & 72.97                     \\ \cline{1-1} \cline{3-8} 
InstructChild       &                             & \textbf{22.93} & \textbf{13.09} & 70.01 & \textbf{70.42} & \textbf{70.21}  & \textbf{73.12} \\ \bottomrule[1pt]
\end{tabular}}
\caption{\label{auto} Main results of baselines and our method. \textbf{Bold}: the maximum value in the column. }
\end{spacing}
\end{table*}

\subsubsection{Performance Comparison}
Table \ref{auto} shows the results of baselines and our InstructChild method on Classic4Children dataset. 
We find that: 
\textbf{i)} The closed-source LLMs, despite their strong performance in many NLP tasks, do not perform well on the child-friendly literary adaptation (CLA) task. 
Specifically, our InstructChild scores higher than these LLMs under all evaluation metrics, e.g. ``+2.17'' and ``+4.93'' on BS-F1 and Red-CN, respectively, compared to GPT-4o, which has over 10 times more parameters.
The reason is that these LLMs tend to generate detailed explanations of the original texts and produce excessively lengthy outputs \cite{causal-vqg,llm-long,l2c}, resulting in suboptimal performance.
\textbf{ii)} There is a significant performance gap between existing text style transfer (TST) models and our InstructChild, which indicates that the CLA task is not simply a matter of text style transfer but also requires the consideration of children's reading preference for effective adaptation.
These TST models cannot capture the child-friendly style for adaptation
and often include words that are too complex for children.  
By integrating characters' personalities and narrative structure, our InstructChild effectively produces text that is both accessible and engaging for children. 
\textbf{iii)} 
Despite our efforts to continually pre-training the Llama models on Chinese instruction datasets, their performance remains considerably distant from the intended reading level.
Specifically, Llama3-8B and Llama2-13B score significantly lower than other models on the Red-CN metric. Notably, they tend to generate specific phrases that closely match the reference text, leading to a higher BS-P value. Since they do not fully cover all the content in the reference text, the BS-R value is low.
Furthermore, QWen2-7B and GLM4-9B also perform worse than our InstructChild, indicating that straightforward LoRA fine-tuning alone for the LLM is insufficient.

\subsubsection{Ablation Study}
The results of ablation experiments are shown in Table \ref{auto}. 
We observe that: 
\textbf{i)} By removing personality information and narrative structure from the integrative instruction, there is a decline in performance on all metrics. It demonstrates that this additional information is crucial for the LLM to produce text adaptation more similar to those written by humans.
Moreover, although InstructChild w/o Per and InstructChild w/o Nat employ refinement, the drop in readability scores (\ie, Red-CN) indicates that initial instruction tuning can encourage the LLM to further generate sentences that align with the children's reading level.
\textbf{ii)} Then we investigate the impact of the refinement with Red-CN metric. Comparing the result of InstructChild and InstructChild w/o Ref, the refinement stage improves performance, particularly on the Red-CN metric. Simultaneously, most traditional natural language generation metrics also improve. It demonstrates that our Red-CN metric is a reasonable measure of children's reading ability in Chinese literary classics.
\textbf{iii)} Finally, we notice a slight performance improvement when applying the lookahead decoding strategy. It suggests that this strategy can further encourage the LLM to generate child-friendly style text with appropriate readability during inference.

\subsubsection{Human Evaluation Results}
To validate the reliability of our human evaluation, 
we separately calculate standard deviations of each human evaluation metric, as shown in Table \ref{human}. The statistical analysis confirms the faithfulness of our evaluation results. 
Table \ref{human} also presents the results of the human evaluation. 
We find that:
\textbf{i)} Our InstructChild achieves competitive results on the \textbf{Flu} metric with the much larger LLM (\ie, GPT-4o) to generate fluent sentences. For the \textbf{CP} metric, our integrative instruction guides the LLM to simplify or omit certain intricate plots. In contrast, GPT-4o often attempts a more comprehensive paraphrase, which results in higher scores. 
\textbf{ii)} The \textbf{CC} metric indicates that our model is more likely to generate adapted texts with rich character descriptions. 
This outcome is attributed to the inclusion of character personality during instruction tuning, making the adapted content more relatable and engaging for children.
\textbf{iii)} The \textbf{NE} metric indicates that our model significantly outperforms baselines by producing a more concise narrative structure aligned with the children's reading level. 
Upon examining the generated outputs, we observe that GPT-4o often provides overly detailed explanations for certain terms, resulting in longer text than the original. This verbosity and inclusion do not align with children's reading preferences.

\begin{table}[]
\centering
\caption{\label{human}Human evaluation results. Each value is presented as $\tau/\rho$, where $\tau$ is the metric value and $\rho$ is the standard deviation. 
\textbf{Bold}: the maximum value. }
\begin{spacing}{1.}
\resizebox{1.\columnwidth}{!}{
\begin{tabular}{l|c|c|c|c}
\toprule[1pt]
\textbf{Method} & \textbf{Flu}   & \textbf{CP}      & \textbf{CC}    & \textbf{NE}      \\ \hline
GPT-4o          & \textbf{1.95}/0.25   & \textbf{1.85}/0.22    & 1.52/0.18           & 0.95/0.12             \\ \hline
StoryTrans     & 0.86/0.23            & 1.12/0.21             & 0.95/0.15           & 0.89/0.17             \\ \hline
GLM4-9B        & 1.72/0.28            & 1.35/0.16             & 1.56/0.19           & 1.25/0.09             \\ \hline
InstructChild  & 1.92/0.19            & 1.69/0.17             & \textbf{1.83}/0.21  & \textbf{1.78}/0.14    \\ \bottomrule[1pt]
\end{tabular}}
\end{spacing}
\end{table}

\subsection{Case Study}
\begin{figure}[!]
  \centering
  \includegraphics[scale=0.283]{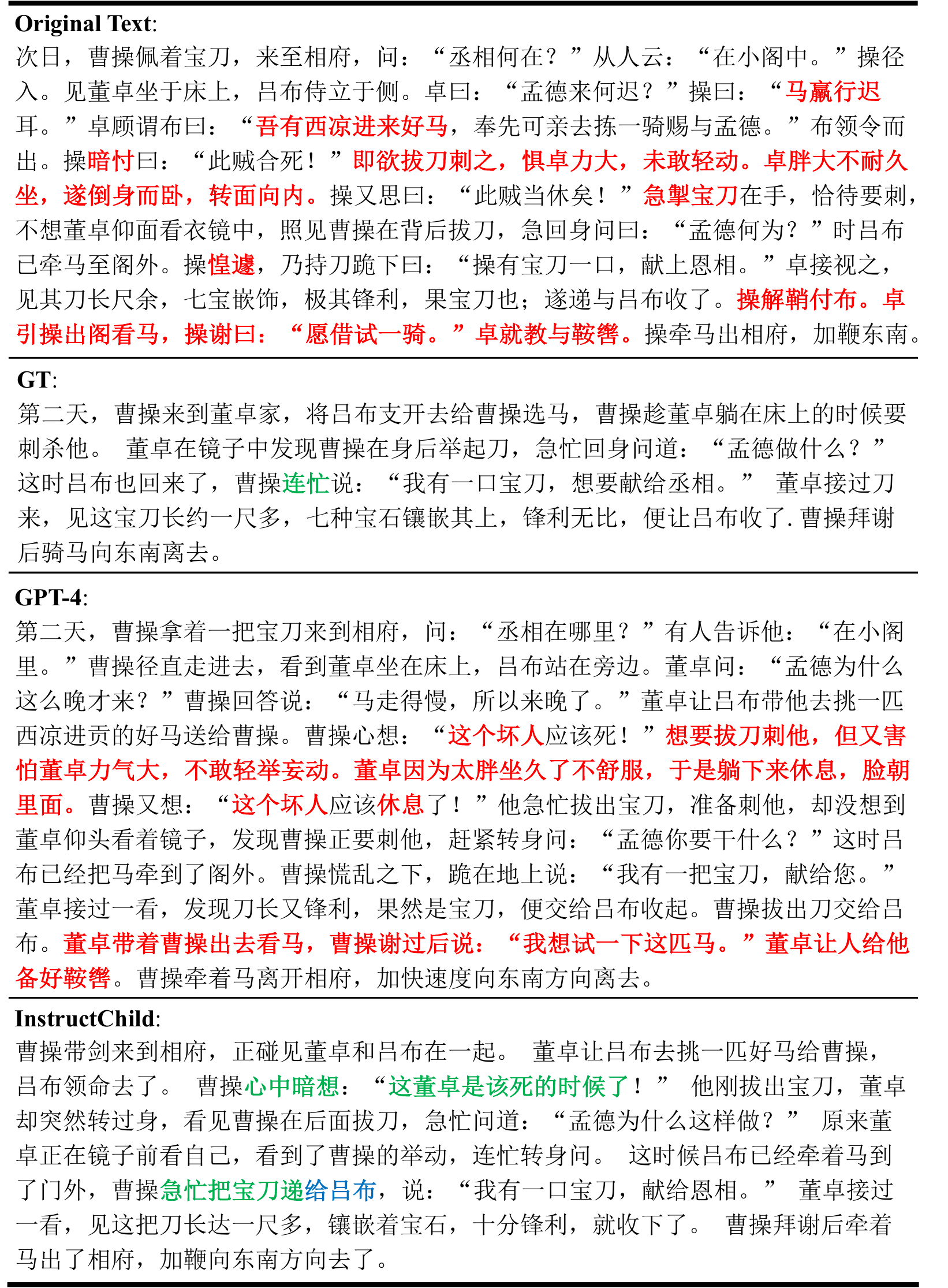}
  \caption{A case study. The \textcolor[RGB]{0, 176, 80}{green text} denotes words that emphasize the characters' personalities. 
  The \textcolor[RGB]{238, 55, 32}{red text} are redundant plots and words with low readability for children.
  The \textcolor[RGB]{0,111,192}{blue text} may cause misunderstanding.}
\label{case}
\end{figure}

Fig. \ref{case} shows the adapted text generated by GPT-4o and our InstructChild.
Specifically, InstructChild produces a more concise narrative that emphasizes key characters' traits.
It describes Cao Cao's determination and urgency while simplifying his inner monologue. In contrast, GPT-4o solely rewrites the original text, resulting in overly lengthy content with complex words. Moreover, the output of GPT-4o is even longer than the original text, further reducing its readability for children.
Although InstructChild performs well on the CLA task, it focuses on the key narrative where Cao Cao's sword ultimately reaches Lv Bu. However, such simplifications can sometimes lead to misunderstandings. 
We believe that the balance between simplifying character relationships and ensuring the clarity of key details is a crucial direction for future research.

\section{Conclusion}
In this paper, we introduce the child-friendly literary adaptation (CLA) task to automatically adapt Chinese literary classics into engaging and age-appropriate text for children.
Moreover, we pinpoint three key children's reading preferences (\ie, vivid character portrayals, concise narrative structures, and appropriate readability) for CLA, even state-of-the-art LLMs struggle to capture these preferences for adaptation.
Our proposed InstructChild explicitly leverages these preferences to guide the LLM in generating child-friendly text for children.
Additionally, we construct the Classic4Children dataset for
a comprehensive evaluation.
Experimental results show that our InstructChild significantly outperforms the existing LLMs.

\section{Limitations}
In this paper, we propose InstructChild to adapt Chinese literary classics into a child-friendly style. 
Although our method significantly enhances the readability of adapted texts for children, the lookahead decoding strategy incurs substantial computational overhead during inference. Consequently, both the number of subsequent tokens $n$ and candidate samples $L$ must be limited. To mitigate these computational costs, a potential solution is to apply distillation techniques to improve decoding efficiency.
Additionally, our CLA task currently focuses on paragraph fragments as training data due to limitations on model input length. In the future, adapting the model to handle the full chapter sequences should be considered.
Moreover, we also believe that the balance between simplifying character relationships and ensuring the clarity of key details is a crucial direction for future research.

\section*{Acknowledgments}
This research is supported by Guangdong Provincial Natural Science Foundation for Outstanding Youth Team Project (2024B1515040010),  the National Natural Science Foundation of China (62076100, 62476097), the Fundamental Research Funds for the Central Universities, South China University of Technology (x2rjD2240100),  Guangdong Provincial Fund for Basic and Applied Basic Research—Regional Joint Fund Project (Key Project) (2023B1515120078).

\bibliography{custom.bib}

\appendix

\newpage
\section{Dataset Construction} \label{app-dataset}

We describe the data collection process based on the Four Great Classical Novels of Chinese literature (\ie, Journey to the West, Romance of the Three Kingdoms, Water Margin, Dream of the Red Chamber). Specifically, the original texts are collected from the publicly available website\footnotemark\footnotetext{https://5000yan.com/}. 
For the children's adapted versions, we choose the widely acclaimed children's reading series\footnotemark\footnotetext{http://www.bph.com.cn/shaoer.html?\_isa=1}.
Following \cite{storytrans}, since there are too many tokens in each chapter of the literary classic, we use paragraph fragments as training samples, with each chapter being divided into multiple fragments.
Considering that a single chapter in the children's version may correspond to multiple chapters in the original version, we first manually align the chapter IDs of the children's edition with those of the original work. This approach effectively narrows down the scope for subsequent annotation, ensuring more accurate correspondence between the versions.
Subsequently, using the text of each paragraph from the children's version as a reference, we manually search for the corresponding text in the original version and constructed them into pairs. Moreover, we filter out paragraphs that lacked a matching counterpart in the original text. This process ensures that the data accurately reflects the intended alignment between the two versions.
After this meticulous process, we obtained 819, 744, 742, and 681 data samples for ``Journey to the West", ``Romance of the Three Kingdoms", ``Water Margin" and ``Dream of the Red Chamber," respectively.
Finally, the collected data are divided into training and testing samples, comprising 2,686 samples for training and 300 samples for testing, with 75 samples from each literary work are randomly selected for testing.

\section{More Experimental Details}

\subsection{Indicators of Readability Metric}
\subsubsection{Adverbs and Conjunctions} 
\label{app-met-1}
Inspired by \cite{child-red}, the appropriate proportion of adverbs and conjunctions in sentences plays a crucial role in the readability of Chinese texts. 
Therefore, we utilize publicly available code\footnotemark\footnotetext{https://pypi.org/project/cntext/} for the calculation.

\subsubsection{Chinese Character Frequency}
\label{app-met-2}
In our analysis of Chinese novel and their texts, we encounter difficulties in calculating word frequency. It arises primarily due to the prevalence of character names and proprietary nouns, which are seldom included in common word lists. As a result, these terms often exhibit anomalously low frequency, even they are simple for children to read.
Consequently, we calculate the frequency of individual characters for each sentence based on the collection \cite{cn-sta}.
Specifically, we use only the top 5000 most frequent Chinese characters from the collection and recalibrate their statistical frequencies for calculation.

\begin{figure}[]
  \centering
  \includegraphics[scale=0.33]{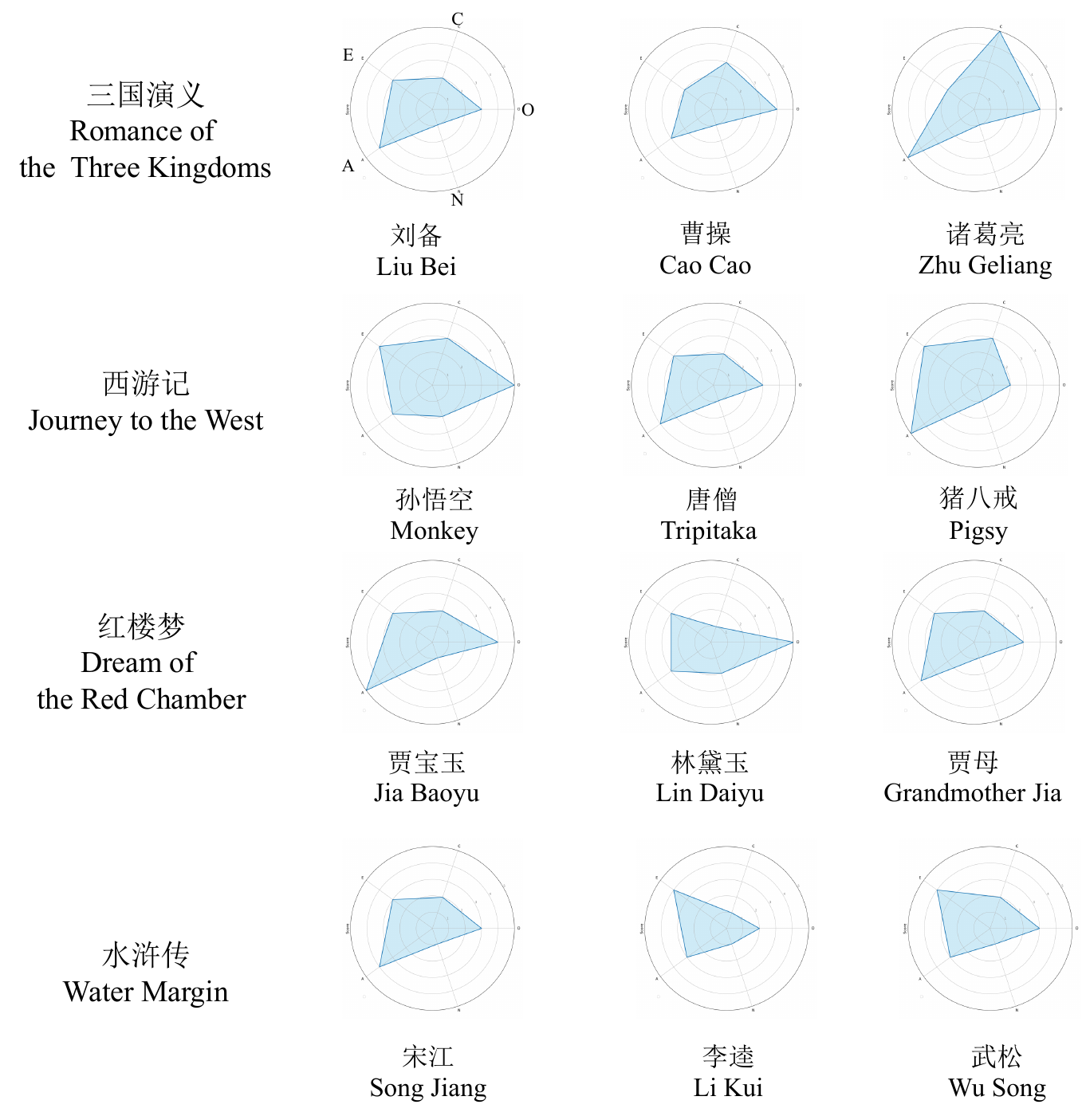}
  \caption{Some cases of personality assessment.} 
  \label{person_case}
\end{figure}

\begin{table}[]
\centering
\caption{\label{hyp_l} Experimental results of different $L$ for token selection. \textbf{Bold}: the maximum value in the column. }
\begin{spacing}{1.1}
\begin{tabular}{l|c|c|c}
\toprule[1pt]
\textbf{$L$} & \textbf{BLEU-2} & \textbf{BS-F1}& \textbf{Red-CN}   \\ \hline
2      & 12.86           & 69.98   & 73.01    \\ \hline
5      & 13.09        & 70.21   & 73.12          \\ \hline
8      & \textbf{13.27}        & \textbf{70.67}            & \textbf{73.15} 
\\ \bottomrule[1pt]
\end{tabular}
\end{spacing}
\end{table}

\subsubsection{Sentence Length} Furthermore, \citet{less-word} suggest that reducing the number of characters contributes to more effective adaptations of literary classics for children. 
We utilize the sentence length of the adapted text as an additional indicator to prevent generating overly verbose text. 
This indicator $r_t$ is also normalized to the range of [0, 1], using its proportional value relative to the original input text, where higher values represent fewer tokens: 
\begin{equation}
    r_t = \max\left(0, 1 - \frac{\mathrm{output\_len}}{\mathrm{input\_len}}\right),
\end{equation}
where $\mathrm{input\_len}$ and $\mathrm{output\_len}$ represent the length of the original text and the generated text, respectively.

\begin{table}[]
\centering
\caption{\label{hyp_n} Experimental results of different $n$ for token selection. \textbf{Bold}: the maximum value in the column. }
\begin{spacing}{1.1}
\begin{tabular}{l|c|c|c}
\toprule[1pt]
\textbf{$n$} & \textbf{BLEU-2} & \textbf{BS-F1}& \textbf{Red-CN}   \\ \hline
10      & 12.94           & 69.91  & 72.80    \\ \hline
20      & 13.09           & 70.21   & 73.12    \\ \hline
50      & \textbf{13.17}          &  \textbf{70.46}   & \textbf{73.64} 
\\ \bottomrule[1pt]
\end{tabular}
\end{spacing}
\end{table}

\begin{table}[]
\centering
\caption{\label{hyp_lambda} Experimental results of different $\lambda$ for the token selection. \textbf{Bold}: the maximum value in the column. }
\begin{spacing}{1.1}
\begin{tabular}{l|c|c|c}
\toprule[1pt]
\textbf{$\lambda$} & \textbf{BLEU-2} & \textbf{BS-F1}& \textbf{Red-CN}   \\ \hline
0.5      & \textbf{13.17}           & 70.10   & 72.42    \\ \hline
1      & 13.09  & \textbf{70.21}   & \textbf{73.12}          \\ \hline
2       & 12.89          & 69.87            & 72.61 
\\ \bottomrule[1pt]
\end{tabular}
\end{spacing}
\end{table}

\subsection{Correlation Comparison}\label{sec-coff}
Following \cite{instructscore}, we report Pearson’s correlation coefficients between the automatic metrics (\ie, BERTScore and reward calculation Red-CN) and human evaluation metrics (\ie, Flu, CP, CC and NE, as defined in the Human Evaluation Criteria section) to assess the effectiveness of our reward calculation. The Core results are shown in Table \ref{coff}, and the p-values of our results are lower than 0.05, which indicates statistical significance.
Specifically, we sample 100 generated instances from the model and invite volunteers with good educational backgrounds in Chinese to provide human evaluation results. The experimental results show that all human evaluation metrics exhibit statistically significant correlations with Red-CN, demonstrating the effectiveness of our reward calculation.

\begin{table}[]
\centering
\caption{\label{coff}Pearson correlation on data samples from Classic4Children dataset.}
\begin{spacing}{1.}
\resizebox{1.\columnwidth}{!}{
\begin{tabular}{l|c|c|c|c}
\toprule[1pt]
\textbf{Metric} & \textbf{Flu} & \textbf{CP} & \textbf{CC} & \textbf{NE} \\ \hline
BERTScore & 23.25 & 15.87 & 24.66 & 18.36 \\ \hline
Red-CN    & \textbf{41.64} & \textbf{34.93} & \textbf{43.39} & \textbf{37.95} \\ \bottomrule[1pt]
\end{tabular}}
\end{spacing}
\end{table}

\subsection{Experiment on Hyperparameter}
We conduct experiments to investigate the sensitivity of hyperparameters during inference, including the number of candidate samples $L$, the number of subsequent tokens $n$ and the weight $\lambda$ for the token selection.
Specifically, 
we explore different parameter $\lambda$ settings, including ($\lambda = 0.5, 1, 2$) in Table \ref{hyp_lambda}, and find that the fluctuations of the results are minimal.
Table \ref{hyp_l} and Table \ref{hyp_n} show the results of different $L$ and $n$, respectively. We observe that the model performance has a slight improvement as $L$ and $n$ increase. However, larger values of $L$ and $n$ are computationally costly, as they require the generation of additional future tokens at each step during inference. Therefore, our InstructChild model employs the parameter setting of $L=5$ and $n=20$ to reduce computational overhead.

\begin{table*}[!]
    \centering
    \small
    \begin{spacing}{1.1}
    \caption{ \label{beft_prompt} The exemplar prompt is used for GPT-4o to assess personality dimension scores of characters based on the Big Five Personality Traits (BFPT) theory.}
    \resizebox{\textwidth}{!}{
    \begin{tabular}{p{\linewidth}}
    \toprule[1pt]
    \textbf{Prompt}: \\
    \begin{CJK*}{UTF8}{gbsn}请根据大五人格特质分析角色：开放性、尽责性、外向性、亲和性、神经质。对西游记中孙悟空的性格进行分析，每个特质分配一个分数（1-5），并为每个分数提供简短的解释（最多 10 个字）。分数越高，表示该特质的存在感越强。请遵循以下格式：
    \end{CJK*} \\
    \begin{CJK*}{UTF8}{gbsn} 经验开放性：2，好奇但保守。\end{CJK*} \\
    \begin{CJK*}{UTF8}{gbsn} 尽责性：1，轻浮不认真。\end{CJK*} \\ 
    \begin{CJK*}{UTF8}{gbsn} 外向性：4，社交活跃。\end{CJK*} \\ 
    \begin{CJK*}{UTF8}{gbsn} 亲和性：3，有些自私。\end{CJK*} \\  
    \begin{CJK*}{UTF8}{gbsn} 神经质：1，易怒冲动。\end{CJK*} \\
    Please analyze the character based on the Big Five personality traits (BFPT): openness to experience, conscientiousness, extraversion, agreeableness, and neuroticism. Analysis of Monkey's character in the Journey to the West. Then assign a score (1-5) for each trait based on the role and provide a brief explanation (max 10 words) for each score. A higher score indicates a stronger presence of that trait. You should use the format as follow: \\
    Openness: 2, curious but reserved. \\
    Conscientiousness: 1, frivolous and not serious. \\
    Extraversion: 4, sociable. \\
    Agreeableness: 3, somewhat selfish. \\
    Neuroticism or emotional stability: 1, irritable and impulsive. \\
\midrule[0.5pt]
\textbf{GPT-4o}: \\ 
 \begin{CJK*}{UTF8}{gbsn} 经验开放性：5，创新冒险。 \end{CJK*} \\
 \begin{CJK*}{UTF8}{gbsn} 尽责性：3，随性自由。\end{CJK*} \\
\begin{CJK*}{UTF8}{gbsn} 外向性：4，活泼好动。\end{CJK*} \\ 
  \begin{CJK*}{UTF8}{gbsn} 亲和性：3，慷慨侠义。\end{CJK*} \\ 
 \begin{CJK*}{UTF8}{gbsn} 神经质：2，冲动易怒。  \end{CJK*} \\
Openness: 5, innovative and adventurous. \\
Conscientiousness: 3, casual and free. \\
Extraversion: 4, lively and active. \\
Agreeableness: 3, generous and chivalrous. \\
Neuroticism: 2, irritable and impulsive. \\
\bottomrule[1pt]
\end{tabular}
    }
    \end{spacing}
\end{table*}

\begin{table*}[!]
    \centering
    \small
    \begin{spacing}{1.1}
    \caption{ \label{ent-rel_prompt} The exemplar prompt is used for GPT-4o to extract the entity-relation triplets as the narrative structure.}
     \resizebox{\textwidth}{!}{
    \begin{tabular}{p{\linewidth}}
    \toprule[1pt]
    \textbf{Prompt}: \\
    \begin{CJK*}{UTF8}{gbsn} 请根据下面文本，抽取故事叙事过程当中的实体和关系，用格式<A, R, B>的形式来进行输出，其中A和B分别是头实体和尾实体，R代表实体之间的关系。输入文本：那妖道：“你是那里和尚？从那里来？到那里去？”快快说明！”三藏道：“我本是唐朝僧人，奉大唐皇帝敕命，前往西方访求经偈 ...  \end{CJK*} \\
    The demon asked, "Where are you from, monk? Where are you headed? Speak quickly!" Tripitaka replied, "I am a monk from the Tang Dynasty, sent by the Emperor of the Tang Empire on a mission to the West to seek sacred scriptures.  ... \\
\midrule[0.5pt]
\textbf{GPT-4o}: \\ 
 \begin{CJK*}{UTF8}{gbsn} <妖道, 询问, 和尚>，<和尚, 来自, 唐朝>，<和尚, 奉命, 大唐皇帝>，<和尚, 前往, 西方>，<和尚, 目的, 访求经偈> ... \end{CJK*} \\
<The demon priest, inquires of, the monk>, <The monk, comes from, the Tang Dynasty>, <The monk, is ordered by, the Emperor of the Great Tang>, <The monk, is heading to, the West>, <The monk, aims to, seek scriptures> ... \\
\bottomrule[1pt]
    \end{tabular}
    }
    \end{spacing}
\end{table*}

\begin{table*}[!]
    \centering
    \small
    \begin{spacing}{1.1}
    \caption{ \label{instruction} The exemplar integrative instruction.}
    \resizebox{\textwidth}{!}{
    \begin{tabular}{p{\linewidth}}
    \toprule[1pt]
    \textbf{Instruction}: \\
    \begin{CJK*}{UTF8}{gbsn} 
    请将以下中国经典名著原文内容改编为适合儿童阅读的版本。你需要依据大五人格理论对角色评估的分数，突出人物的性格特征。并利用实体-关系三元组来构建故事的叙事框架，适当简化或省略一些复杂的叙事情节。
    \end{CJK*} \\
    \# \begin{CJK*}{UTF8}{gbsn} 人物性格: \end{CJK*} \\
    \begin{CJK*}{UTF8}{gbsn} 孙悟空：[经验开放性：5，创新冒险。尽责性：3，随性自由。外向性：4，活泼好动。亲和性：3，慷慨侠义。神经质：2，冲动易怒 ...] \end{CJK*} \\
    \# \begin{CJK*}{UTF8}{gbsn} 实体关系三元组：\end{CJK*} \\
    \begin{CJK*}{UTF8}{gbsn} <道士，在，滚油锅里>，...，<国王，扑在，御案>。 \end{CJK*} \\
    \# \begin{CJK*}{UTF8}{gbsn} 原文内容：\end{CJK*} \\
    \begin{CJK*}{UTF8}{gbsn} 三藏、八戒、沙僧立在殿前，见那道士在滚油锅里打挣 ...  \end{CJK*} \\
    Please adapt the following Chinese classics into a version suitable for children. You need to highlight the character's personality traits based on the evaluation scores. Use entity-relationship triples to construct the story's narrative framework, and appropriately simplify or omit some complex narrative plots. \\
    \# Character: \\ 
    Monkey: Openness: 5, innovative and adventurous. Conscientiousness: 3, casual and free. Extraversion: 4, lively and active. Agreeableness: 3, generous and chivalrous. Neuroticism: 2, impulsive and irritable. ... \\
    \# Entity-relationship triples: \\ 
    <Taoist priest, in, boiling oil pot>, ..., <King, fell on, the imperial case>. \\
    \# Original content: \\ 
    Tripitaka, Pigsy, and Friar Sand stood in front of the hall and saw the Taoist priest struggling in the boiling oil pot ... \\
    \bottomrule[1pt]
    \end{tabular}
    }
    \end{spacing}
\end{table*}

\begin{table*}[!]
    \centering
    \small
    \begin{spacing}{1.1}
    \caption{ \label{prompt_llms} The prompt template for GPT-4o, GLM-4, Qwen2.5 baselines.}
     \resizebox{\textwidth}{!}{
    \begin{tabular}{p{\linewidth}}
    \toprule[1pt]
    \textbf{Prompt}: \\
    \begin{CJK*}{UTF8}{gbsn} 请将以下中国经典名著的原文改写为儿童易懂的版本，参考以下示例格式：\end{CJK*} \\
\# \begin{CJK*}{UTF8}{gbsn} 示例：\end{CJK*} \\
\#\# \begin{CJK*}{UTF8}{gbsn} 原文内容：\end{CJK*} \\
\begin{CJK*}{UTF8}{gbsn} 孙悟空在旁闻讲，喜得他抓耳挠腮，眉花眼笑。忍不住手之舞之，足之蹈之。忽被祖师看见，叫孙悟空道：“你在班中，怎么颠狂跃舞，不听我讲？”悟空道：“弟子诚心听讲，听到老师父妙音处，喜不自胜，故不觉作此踊跃之状。望师父恕罪！”祖师道：“你既识妙音，我且问你，你到洞中多少时了？”悟空道：“弟子本来懵懂，不知多少时节。只记得灶下无火，常去山后打柴，见一山好桃树，我在那里吃了七次饱桃矣。”祖师道：“那山唤名烂桃山。你既吃七次，想是七年了。” \end{CJK*} \\
\#\# \begin{CJK*}{UTF8}{gbsn} 输出： \end{CJK*} \\
\begin{CJK*}{UTF8}{gbsn} 悟空听到精彩的地方，高兴得抓耳挠腮。祖师很生气，问他：“你怎么在下面疯疯癫癫的，不听我讲？”悟空说：“弟子在认真地听，听到精妙的地方，喜不自禁，希望师父原谅！”祖师说：“你到洞中有多长时间了？”悟空说：“弟子不知，只记得经常去山后打柴，在那里吃了七次饱桃了。”祖师说：“那应当是七年了。” \end{CJK*} \\
\#\# \begin{CJK*}{UTF8}{gbsn} 原文内容： \end{CJK*} \\ 

\begin{CJK*}{UTF8}{gbsn} \{原文内容\} \end{CJK*} \\
Please rewrite the original text of the following Chinese classics into a version that is easy for children to understand, referring to the following sample format: \\
\# Example: \\
\#\# Original content: \\ 
Monkey was listening to the lecture and was so happy that he scratched his ears and cheeks, smiling. He couldn't help dancing with his hands and feet. Suddenly, the master saw him and called Monkey, saying, "Why are you dancing wildly in the class instead of listening to me?" Monkey said, "I listened to the lecture sincerely. I was overwhelmed with joy when I heard the wonderful sound of the master, so I jumped up and down unconsciously. I hope the master will forgive me!" The master said, "Since you know the wonderful sound, let me ask you, how long have you been in the cave?" Monkey said, "I was ignorant and didn't know how long it had been. I only remember that there was no fire under the stove, so I often went to the back of the mountain to collect firewood. I saw a mountain of beautiful peach trees, and I ate peaches there seven times." The master said, "That mountain is called Rotten Peach Mountain. Since you have eaten seven times, it must be seven years." \\
\#\# Output: \\ 
Monkey was so happy to hear the wonderful part that he scratched his cheeks. The master was very angry and asked him, "Why are you acting crazy down there and not listening to me?" Monkey said, "I am listening carefully. I am so happy to hear the wonderful part. I hope you will forgive me!" The master asked, "How long have you been in the cave?" Monkey said, "I don't know. I just remember that I often went to the back of the mountain to collect firewood and ate peaches seven times there." The master said, "That should be seven years." \\
\#\# Original content: \\
\{Original content\} \\
    \bottomrule[1pt]
    \end{tabular}
    }
    \end{spacing}
\end{table*}

\begin{table*}[!]
    \centering
    \small
    \begin{spacing}{1.1}
    \caption{ \label{prompt_EAPMT} The prompt template for EAPMT baseline.}
     \resizebox{\textwidth}{!}{
    \begin{tabular}{p{\linewidth}}
    \toprule[1pt]
    \textbf{Prompt}: \\
    \begin{CJK*}{UTF8}{gbsn} 请对以下输入内容生成详细的帮助儿童理解的解释，并结合解释请将以下中国经典名著的原文改写为儿童易懂的版本，参考以下示例格式:\end{CJK*} \\
    
\# \begin{CJK*}{UTF8}{gbsn} 示例：\end{CJK*} \\
\#\# \begin{CJK*}{UTF8}{gbsn} 原文内容：\end{CJK*} \\
\begin{CJK*}{UTF8}{gbsn} 孙悟空在旁闻讲，喜得他抓耳挠腮，眉花眼笑。忍不住手之舞之，足之蹈之。忽被祖师看见，叫孙悟空道：“你在班中，怎么颠狂跃舞，不听我讲？”悟空道：“弟子诚心听讲，听到老师父妙音处，喜不自胜，故不觉作此踊跃之状。望师父恕罪！”祖师道：“你既识妙音，我且问你，你到洞中多少时了？”悟空道：“弟子本来懵懂，不知多少时节。只记得灶下无火，常去山后打柴，见一山好桃树，我在那里吃了七次饱桃矣。”祖师道：“那山唤名烂桃山。你既吃七次，想是七年了。” \end{CJK*} \\
\#\# \begin{CJK*}{UTF8}{gbsn} 解释：\end{CJK*} \\
\begin{CJK*}{UTF8}{gbsn} 故事里，孙悟空表现出对学习的喜爱，他听课时因为兴奋而不自觉地跳起来，表现了他很喜欢学习，并且心中充满了好奇和快乐。师父也通过询问，指出了他在学习的进步，暗示了孙悟空在师父的教导下已经学到了很多东西。这个情节让我们看到了孙悟空的活泼和热爱学习的精神，同时也反映了师徒之间的有趣互动。\end{CJK*} \\

\#\# \begin{CJK*}{UTF8}{gbsn} 输出：\end{CJK*} \\ 
\begin{CJK*}{UTF8}{gbsn} 孙悟空在一旁听着师父讲课，听得特别开心，忍不住搔搔头、笑得眯起了眼睛，心里美滋滋的。高兴得他开始手舞足蹈，开心地跳了起来。可是他正跳着呢，突然被师父看到了。师父说：“孙悟空，你在课堂上怎么这样乱跳乱舞，不认真听课呢？”孙悟空连忙解释说：“师父，我真的在认真听呢！只是因为您的讲课太精彩了，让我忍不住开心地跳了起来。希望师父不要怪我！”师父听了，微笑着说：“既然你听得这么认真，那我来问你一个问题。你记得自己在这里待了多久了吗？”孙悟空挠挠头说：“师父，我不知道具体的时间。只记得这里的厨房没火，我经常去山后砍柴。那儿有一棵很好的桃树，我在那里吃了七次桃子，都吃得饱饱的。”师父笑了笑，说：“那座山叫‘烂桃山’。你已经吃了七次桃子，说明你在这里学习了七年啦！”\end{CJK*} \\ 
\#\# \begin{CJK*}{UTF8}{gbsn} 原文内容：\end{CJK*} \\
\begin{CJK*}{UTF8}{gbsn} \{原文内容\} \end{CJK*} \\
Please generate detailed explanations for the following input content to help children understand it, and based on the explanations, please rewrite the original text of the following Chinese classics into a version that is easy for children to understand: \\
\# Example: \\
\#\# Original content: \\ 
Monkey was listening to the lecture and was so happy that he scratched his ears and cheeks, smiling. He couldn't help dancing with his hands and feet. Suddenly, the master saw him and called Monkey, saying, "Why are you dancing wildly in the class instead of listening to me?" Monkey said, "I listened to the lecture sincerely. I was overwhelmed with joy when I heard the wonderful sound of the master, so I jumped up and down unconsciously. I hope the master will forgive me!" The master said, "Since you know the wonderful sound, let me ask you, how long have you been in the cave?" Monkey said, "I was ignorant and didn't know how long it had been. I only remember that there was no fire under the stove, so I often went to the back of the mountain to collect firewood. I saw a mountain of beautiful peach trees, and I ate peaches there seven times." The master said, "That mountain is called Rotten Peach Mountain. Since you have eaten seven times, it must be seven years."\\
\#\# Explanation:\\
In the story, Monkey showed his love for learning. He jumped up unconsciously because of excitement during the class, which showed that he liked learning very much and was full of curiosity and joy. The master also pointed out his progress in learning through questioning, implying that Monkey had learned a lot under the guidance of the master. This plot allows us to see Monkey's liveliness and love of learning, and also reflects the interesting interaction between the master and the apprentice.\\
\#\# Output: \\
Sun Wukong was listening to the master's lecture. He was so happy that he couldn't help scratching his head and smiling until his eyes narrowed. He was so happy that he began to dance and jump up happily. But when he was jumping, he was suddenly seen by the master. The master said: "Sun Wukong, why are you dancing around in class and not listening carefully?" Sun Wukong quickly explained: "Master, I am really listening carefully! It's just that your lecture is so wonderful that I can't help but jump up happily. I hope Master won't blame me!" The master smiled and said: "Since you listen so carefully, let me ask you a question. Do you remember how long you have been here?" Sun Wukong scratched his head and said: "Master, I don't know the exact time. I only remember that there is no fire in the kitchen here, so I often go to the back of the mountain to chop firewood. There is a very good peach tree there. I ate peaches there seven times and was full every time." The master smiled and said: "That mountain is called 'Rotten Peach Mountain'. You have eaten peaches seven times, which means you have been studying here for seven years!"\\ 
\#\# Original content: \\
\{Original content\} \\
    \bottomrule[1pt]
    \end{tabular}
    }
    \end{spacing}
\end{table*}

\end{document}